\pdfoutput=1

\documentclass[11pt]{article}
\usepackage[final]{acl}

\usepackage{times}
\usepackage{latexsym}

\usepackage[T1]{fontenc}

\usepackage[utf8]{inputenc}

\usepackage{microtype}

\usepackage{inconsolata}

\usepackage{graphicx}
\usepackage{xspace}
\usepackage{amsmath}

\usepackage{amssymb}
\usepackage{booktabs}
\usepackage{cleveref}
\usepackage{multirow}

\usepackage{acronym}

\newcommand{\newpara}[1]{\vspace{0.2cm}\noindent \textbf{#1}}
\newcommand{\method}{\emph{Slam}\xspace}
\newcommand{\slm}{SLM\xspace}
\newcommand{\slms}{SLMs\xspace}
\newcommand{\ssc}{\textsc{sSC}\xspace}
\newcommand{\tsc}{\textsc{tSC}\xspace}

\usepackage{acronym}

\acrodef{LM}{Language Model}
\acrodef{LLM}{Large Language Model}
\acrodef{LMs}{Language Models}
\acrodef{LLMs}{Large Language Models}
\acrodef{uLM}{unit Language Model}
\acrodef{SOTA}{State-Of-The-Art}
\acrodef{GSLM}{Generative Spoken Language Modeling}
\acrodef{TTS}{Text-to-Speech}
\acrodef{WER}{Word-Error-Rate}
\acrodef{ASR}{Automatic Speech Recognition}
\acrodef{SSL}{Self-Supervised Learning}
\acrodef{NLP}{Natural Language Processing}
\acrodef{PPL}{Perplexity}

\usepackage{booktabs}
\usepackage{hyperref}
\usepackage{pifont}
\newcommand{\cmark}{\ding{51}}%
\newcommand{\xmark}{\ding{55}}%

\title{\textit{Slamming}: Training a Speech Language Model on One GPU in a Day}

\author{Gallil Maimon*, Avishai Elmakies*, Yossi Adi \\
        *Equal Contribution \\ The Hebrew University of Jerusalem \\ \texttt{gallil.maimon@mail.huji.ac.il}}

\begin{document}
\maketitle
\begin{abstract}
We introduce \method, a recipe for training high-quality Speech Language Models (SLMs) on a single academic GPU in 24 hours. We do so through empirical analysis of model initialisation and architecture, synthetic training data, preference optimisation with synthetic data and tweaking all other components. We empirically demonstrate that this training recipe also scales well with more compute getting results on par with leading \slms in a fraction of the compute cost. We hope these insights will make \slm training and research more accessible. In the context of \slm scaling laws, our results far outperform predicted compute optimal performance, giving an optimistic view to \slm feasibility. See code, data, models, samples - \href{https://pages.cs.huji.ac.il/adiyoss-lab/slamming}{https://pages.cs.huji.ac.il/adiyoss-lab/slamming}.
\end{abstract}

\section{Introduction}

Speech Language Models (\slms) have gained significant interest from researchers~\cite{peng2024survey, cui2024recent, ji2024wavchat, latif2023sparks}, demonstrating remarkable performance in traditional speech tasks~\cite{valle, elmakies2025unsupervisedspeechsegmentationgeneral}, diverse generative applications~\cite{yang2023uniaudio, yang2024uniaudio}, and reasoning over speech and audio signals~\cite{salmonn, qwen_audio}.

\slms can generally be classified into two main categories: (i) generative speech \ac{LMs} (which can also incorporate text) and (ii) speech-aware \ac{LMs}. The first category follows a similar pre-training approach to text-based \ac{LLMs}, directly maximising the likelihood of speech considering both input and output, typically by representing audio as a sequence of discrete tokens. The second category consists of pre-trained text \ac{LMs} adapted to process speech inputs. In this work, we focus on the first.

\begin{figure}[t]
  \includegraphics[width=\columnwidth]{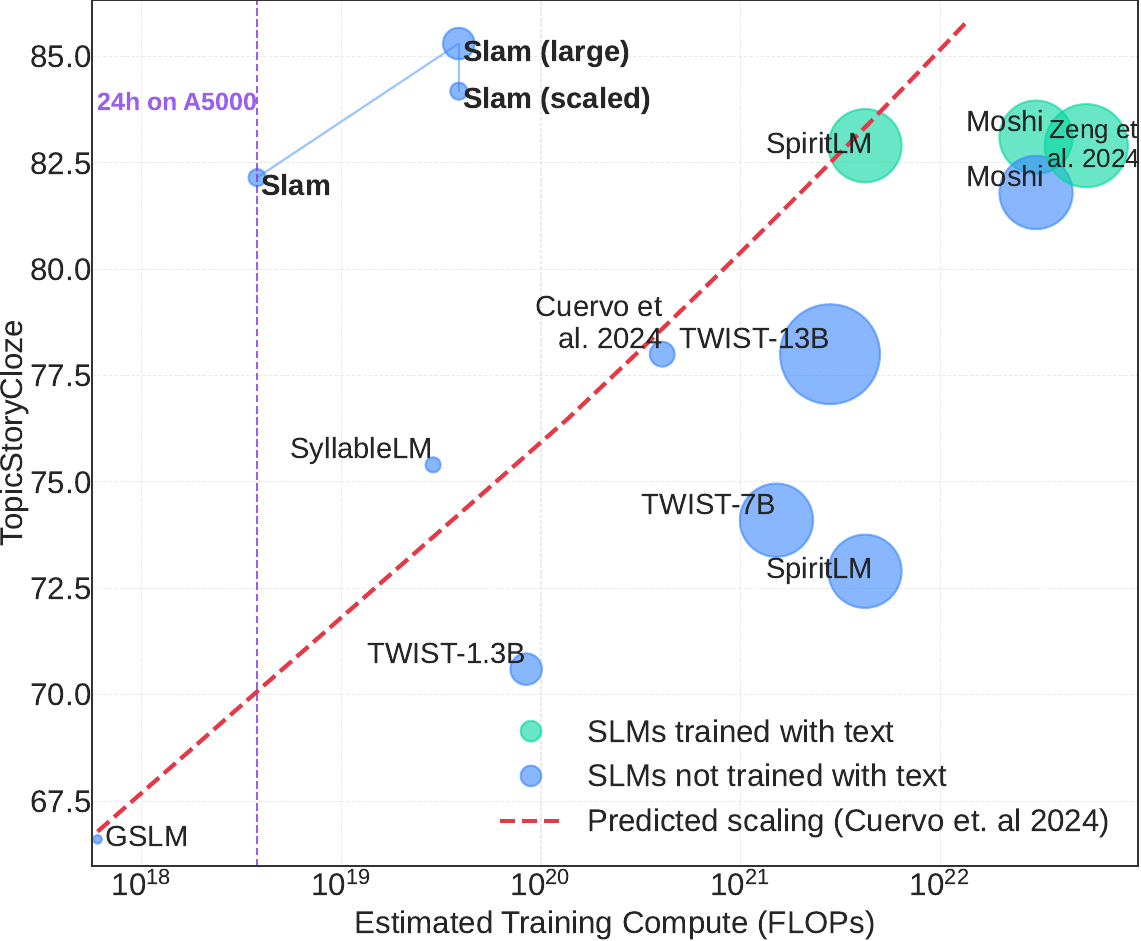}
  \caption{Comparing Topic-StoryCloze performance of different \slms as a function of training compute. Model size is indicated by the size of the circle.}
  \label{fig:teaser}
\end{figure}

Training high-quality \slms can be highly resource intensive~\cite{twist, cuervo2024scaling, scaling_interleaving, spiritlm, defossez2024moshi}. For example, ~\citet{spiritlm} trained their \slm on approximately $570k$ hours of speech data, while ~\citet{defossez2024moshi} utilised around $7M$ hours. Additionally, ~\citet{cuervo2024scaling} proposed \slm scaling laws, suggesting that training high-quality \slms requires $\sim3X$ more data compared to text-based counterparts. These computational demands restrict the required fundamental research aimed at enhancing \slms, such as advancements in speech tokenisation, efficient acoustic modelling, etc.

In the \ac{NLP} community, numerous studies have investigated efficient model training techniques, including masked language models such as Cramming~\citep{geiping2023cramming} and ModernBERT~\citep{warner2024modernbert}, along with next-token prediction LLMs such as MobileLLM~\citep{mobilellm}. These methods include implementation efficiencies, architectural improvements, data selection strategies, and enhancements to the overall training pipeline. 

Inspired by Cramming~\cite{geiping2023cramming} in text, we investigate compute-limited \slm training, which we term \emph{Slamming}. We pose the question: \emph{Is it possible to train high-quality \slms using a single GPU within 24 hours?} For that, we conduct an extensive empirical analysis exploring how different training components influence performance. From this, we derive a training recipe that maximises model performance within a fixed compute budget. Specifically, we investigate the impact of model initialisation and architecture, various optimisers and learning rate schedulers, data selection strategies - including the role of synthetic data, text-interleaving and preference optimisation. 

We believe that developing these training strategies and proving their feasibility will empower the speech and audio research community to advance \slms beyond the scope of large, well-funded academic and industrial labs. Figure~\ref{fig:teaser} illustrates the performance of various \slms relative to their training compute budget, with circle sizes representing the size of the models. Furthermore, we compare our results with the scaling performance predicted from \citet{cuervo2024scaling}. Although the authors present a somewhat pessimistic view of the computational resources needed to train high-quality \slms, we empirically show that reality is more promising, demonstrating that it is possible to significantly exceed the predicted performance per unit of compute. We encourage the community to refine and expand scaling laws specifically tailored for \slm training across various settings.

\newpara{Our Main Contributions are:} 
\begin{enumerate}
    \item We introduce \method, a training recipe for efficiently training high-quality \slms using a single $A5000$ GPU within $24$ hours.
    \item We carry out extensive experiments exploring model initialisation and architecture, optimisation, data collection and generation, and training objectives (i.e., preference optimisation and text-speech interleaving), providing insights into the impact of each component on model performance.
    \item Building on these insights, we scale the compute budget to two $A100$ GPUs for $48$ hours and demonstrate that our model achieves performance on par with state-of-the-art models that require substantially more compute.
\end{enumerate}
We open-source all code, models, training recipes, and synthetic datasets.
\section{Related Work}

\newpara{Efficient Training.} Enhancing the efficiency of neural network training has been extensively studied~\citep{shen2023efficient}. \citet{hajimolahoseini2023swiftlearn, wang2024greats} examined the impact of data selection on \ac{LLM} training and introduced efficient data selection methods. \citet{muhamed2024grass} proposed using structured sparse gradients to enhance compute efficiency in \ac{LLM} training, while \citet{rawat2024little} explored the potential of leveraging smaller language models to improve the training efficiency of larger \ac{LLM}s. \citet{lv2024scalable} investigated the use of low-dimensional projections for attention parameters to enhance training efficiency. Meanwhile, \citet{neiterman2024layerdropback} proposed applying LayerDrop as a technique to optimise neural network training.

More closely related to our work, \citet{li2023flm} propose a training strategy for developing \ac{LLM}s within a $100k\$$ budget. \citet{warner2024modernbert} introduce ModernBERT, an efficient training pipeline for optimising BERT models, while \citet{izsak2021train} outline a method for training a BERT model in $24$ hours using $8$ GPUs. The most relevant work to ours is Cramming \cite{geiping2023cramming}, where the authors conduct an in-depth analysis of masked LM training on a single GPU in one day.

While these studies offer valuable insights, they primarily focus on training text models, such as \ac{LLM}s and masked LMs. In the speech domain, similar research has been conducted on self-supervised representation models~\citep{liu2024efficient}, but not on \slms. In this work, we address this gap by focusing on efficient \slm training.

\newpara{Generative Speech Language Models} were explored under various setups~\citep{gslm, kharitonov2021text}.  
\citet{gslm} were the first to show how raw, uncurated speech data can be leveraged into building a Generative Speech Language Model (GSLM). Next,~\citet{borsos2023audiolm} proposed a cascade version using both coarse and fine speech tokens. Such a modelling framework opened up a new and promising research direction for processing and modelling spoken data, such as speech resynthesis~\citep{polyak2021speech}, speaking style conversion~\citep{kreuk2021textless,maimon2023dissc}, dialogue modelling~\cite{nguyen2022generative}, speech-to-speech translation~\citep{popuri2022enhanced, peng2024mslm}, etc. 
\citet{nachmani2023spoken} proposed augmenting a text \ac{LM} with continuous speech data to improve spoken question-answering tasks. Recently, \citet{park2024long} proposed \slm based on state-space models~\citep{gu2021efficiently} to further push long context-efficient modelling, while ~\citet{lin2024alignslm} proposed to fine-tune \slms using direct preference optimisation~\citep{rafailov2024dpo} obtained from text LLM rankings. 

Similar to text \ac{LLMs}, training \slms often demands large-scale datasets. For instance, Moshi~\cite{defossez2024moshi} was trained on $7$ million hours of speech data, SpiritLM~\cite{spiritlm} utilized $560k$ hours, and TWIST~\cite{twist} was trained on approximately $150k$. Recently, \citet{cuervo2024scaling} introduced the first scaling laws for \slms, suggesting that achieving comparable performance to text LMs requires three times more tokens. In this work, we focus on reducing the computational demands while maintaining performance comparable to leading \slms. 
\section{Setup}
In this study, we explore decoder-only generative \slms, which aim at maximising the likelihood of speech samples represented as discrete tokens. We examine both purely speech-based \slms trained on speech tokens and joint speech-text \slms using interleaving strategies~\citep{spiritlm}. Similarly to ~\citet{twist, gslm}, we obtain speech tokens by quantising continuous latent representations of a self-supervised speech representation model using the k-means algorithm, often known as \emph{semantic tokens}. Specifically, we utilise a multilingual HuBERT~\citep{hubert} model running at $25$ Hz, as employed in ~\citet{twist}. We then train \slms by minimising the negative log-likelihood of the input segments.

Unless mentioned otherwise, all \slms are trained using \textbf{a single $A5000$ GPU ($24GB$ VRAM)} along with $16$ CPU cores for $24$ hours. We deliberately focus on this constrained compute budget, assuming that most academic labs can access similar resources, thereby ensuring the accessibility of our research. The training data is pre-processed, i.e. extracting HuBERT units and dividing data into chunks, and stored prior to model training. As a result, this pre-processing time is excluded from the compute budget. This approach, aligned with \citet{geiping2023cramming}, is practical since many research experiments utilise the same pre-processed data. We additionally do not count the time for running validation and visualisations as they are not used as part of the optimisation pipeline and only used for demonstration purposes.

\label{para:metrics}
\newpara{Evaluation Metrics.} We assess all \slms using five distinct evaluation metrics. The first three are based on likelihood evaluation, while the fourth and fifth are generative metrics. For likelihood based modelling we consider sBLIMP~\cite{dunbar2021zero}, \emph{Spoken Story-Cloze} (\ssc)), and \emph{Topic Story-Cloze} (\tsc)~\cite{twist}. For modelling-likelihood metrics, we evaluate the likelihood assigned by the \slms to pairs of speech utterances, consisting of a positive example and a distractor. We calculate the percent of pairs in which the \slm assigns higher likelihood to the positive sample. sBLIMP focuses on grammatical abilities thus the negative is ungrammatical version of the positive. \ssc and \tsc focus on semantic modelling abilities. In \ssc, the distractor suffix is taken from the original textual StoryCloze dataset~\citep{mostafazadeh2016corpus}, allowing to assess fine-grained semantic speech understanding. In \tsc, however, the distractor suffix is drawn from a different topic, enabling us to evaluate the model’s ability to understand the overall semantic concept. 

To assess the generative abilities of \slms, we compute \emph{generative perplexity} (GenPPL). Following the approach of~\citet{gslm, twist}, we provide the \slm with a short speech prompt and generate speech tokens continuation. We use unit-vocoder with duration prediction to convert the tokens into speech~\citep{polyak2021speech, twist}. The generated speech is then transcribed, and its \ac{PPL} is evaluated using a pre-trained text \ac{LLM}. To minimise the impact of token repetition on \ac{PPL} measurements, we ground the generated text using diversity metrics derived from the auto-BLEU score~\citep{gslm}. Similarly to ~\citet{lin2024alignslm} we use bigram auto-BLEU. In other words, we ensure that all models achieve similar auto-BLEU scores, allowing for a fair comparison of \ac{PPL}. Specifically, we transcribe speech segments using Whisper-large-v$3$-turbo model~\citep{radford2023robust} and measure \ac{PPL} using Llama-$3.2$-$1$B model~\citep{grattafiori2024llama3herdmodels}. We calculate GenPPL on correct samples from the Spoken Story-Cloze dataset. 

Finally, for our final models, we also compute \emph{GPTScore}. Given a speech prompt and a generated continuation, we transcribe both and use GPT-4o to judge the quality of the continuation given the prompt, on a scale of 1 to 5. We follow the same setup and prompt as \citet{lin2024alignslm} for the metric. We use this metric as the final form of evaluation, as it is the most costly to run.

\newpara{Software Efficiency.} To maximise performance within $24$ hours of model training, we leverage multiple efficient implementations. Through extensive performance testing, we found that using bfloat$16$ ~\cite{kalamkar2019study} alongside FlashAttention$2$ \citep{dao2023flashattention2fasterattentionbetter} and data packing provided the most efficient compute performance in our setup. We also experimented with model compilation using \texttt{torch.compile} \citep{ansel2024pytorch}, but it lacked native compatibility with FlashAttention$2$ at the time of our study, and its performance without FlashAttention$2$ was subpar. Future work could investigate this further with more efficient attention implementations~\cite{FA3, li2024flexattention}.

To enable rapid and scalable experimentation, we developed a specialised library for \slm training that supports various model architectures, training objectives, and evaluation metrics. It accommodates TWIST-style training, text-speech interleaving, preference optimisation, etc. We open-source this package along with all model weights and training recipes, aiming to empower the community to further explore \slms.
\section{Investigations}
With this setup, we systematically analyse and ablate each component of the training pipeline, ultimately refining an optimised cook-book for training \slms. We specifically examine the influence of model family, initialisation, size, and architectural choices (e.g., dropout, positional embedding, etc.). We analyse optimisation parameters and data characteristics. Lastly, we explore alternative training objectives beyond standard next-token prediction, including speech-text interleaving and direct preference optimisation using synthetic data. 

\subsection{Model \& Optimisation}
\newpara{Hyper-parameters.} Unless specified otherwise, we use a context length of $512$ tokens and an effective batch size of $256$, employing gradient accumulation when necessary, as preliminary results indicated this configuration yields the best overall performance. We set the peak learning rate to $1e-3$ to enhance training speed and use a warmup period of $1\%$ of the total training steps, as this proved more effective than the fixed $100$-step warmup used in the original TWIST. To improve training stability, particularly with large learning rates, we apply gradient normalisation with a norm of $0.5$ at no additional cost, following~\citet{geiping2023cramming}. Unless modified later in our investigation, we use an inverse-square root scheduler and the AdamW optimiser~\citep{loshchilov2017decoupled}. Likewise, this sub-section uses the common Libri-Speech And Libri-Light datasets for training, until further investigated in Section \ref{sec:data}.

\begin{figure}[t!]
  \includegraphics[width=\columnwidth]{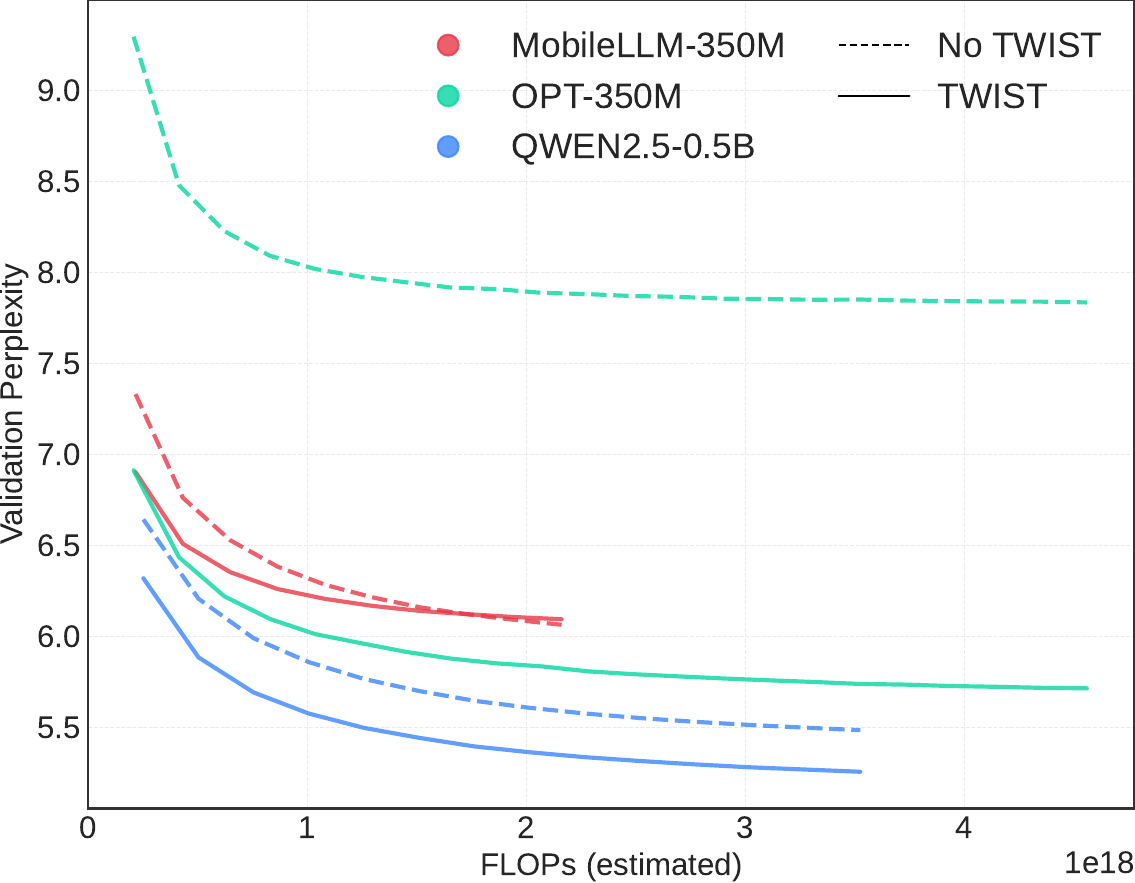}
  \caption{Comparing validation PPL of different models of similar parameter count, with and without TWIST initialisation.}
  \label{fig:initialisation}
\end{figure}

\newpara{Initialisation.} \citet{twist} empirically demonstrated that initialising \slms with pre-trained text \ac{LM}s can enhance convergence speed and improve model performance. We examine the effect of this initialisation within our setup across different model types. To do so, we train multiple models, both with and without TWIST initialisation, while staying within our compute budget. As shown in Figure~\ref{fig:initialisation}, TWIST initialisation benefits all evaluated models at the beginning of training, though its overall impact by the end varies. Notice, the x-axis in Figure~\ref{fig:initialisation} represents theoretical FLOPs, calculated as $6 * N_{\mathrm{params}} * D_{\mathrm{tokens}}$ following~\citet{hoffmann2022training}. However, due to variations in model architecture and implementation, practical efficiency differs, leading to varying amounts of compute processed within $24$ hours. 

Results suggest that benefits of TWIST initialisation can be substantial, especially for top-performing models like Qwen$2.5$. As a result, we prioritise investigations based on existing pre-trained text \ac{LM}s. Interestingly, the results in Figure~\ref{fig:initialisation} demonstrate that Qwen$2.5$ outperforms other models even without TWIST initialisation, perhaps suggesting that their architectural design choices or size might also provide some benefit.

\begin{figure}[t!]
  \includegraphics[width=\columnwidth]{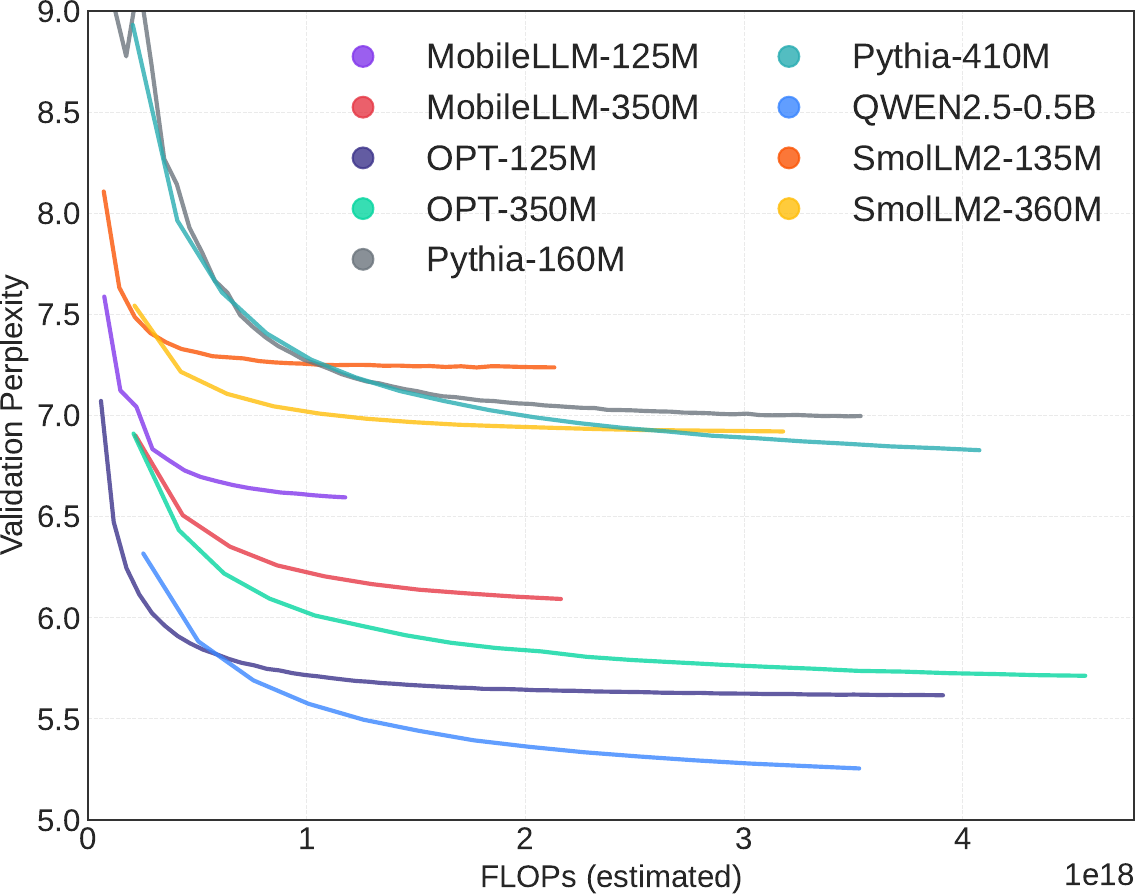}
  \caption{Comparing PPL of different models under TWIST initialisation.}
  \label{fig:model_choice}
\end{figure}

\newpara{Optimal Model Size \& Family.} \citet{cuervo2024scaling} conducted a scaling analysis on GSLM-style \slms, estimating the optimal model size and token count for a compute-efficient model. However, using a text LM initialisation might impact these findings. As we observe, TWIST initialisation greatly impact model performance, suggesting that prioritising larger models may be more effective than simply increasing the dataset size. Additionally, various model families gain different advantages from TWIST initialisation; for example, Qwen$2.5$ models show significantly better performance compared to OPT models. In Figure~\ref{fig:model_choice}, we compare the results under the pre-defined compute budget within model families\footnote{We use the text LM original names for clarity, but note that the actual size will be notably smaller due to reduced vocabulary size, e.g Qwen$2.5$-$0.5$B has $358$M parameters. Full model sizes can be found in Appendix \ref{sec:model_sizes}.}. We note that the best model sizes for MobileLLM \cite{mobilellm}, SmolLM$2$ \cite{smollm2} and Pythia \cite{pythia} are $\sim300M$ parameters, while for OPT the best is $125$M. According to \citet{cuervo2024scaling}, the estimated optimal model size is approximately $66$M parameters. However, the best-performing model, Qwen$2.5$, is significantly larger. Since there are no smaller models in this family, it is difficult to determine whether this deviation is due to the quality of the initialisation or other factors. Moving forward, we proceed with both OPT-$125$M and Qwen$2.5$-$0.5$B.

\newpara{Dropout.} The original OPT models includes dropout to mitigate overfitting. Although dropout is beneficial for regularisation, it effectively decreases the number of gradient updates per parameter without shortening the update-step wall time. Hence, reduces the number of parameter updates per second. Following \citet{geiping2023cramming}, we experiment with removing dropout and observed improved performance in our setup.

\newpara{Positional Encoding.} Transformers rely on positional encoding to capture the order of input tokens. Many modern LMs, including the Qwen models, use Rotary Position Embedding~\citep{su2023roformerenhancedtransformerrotary}. This method uses a hyperparameter, $\theta$, to control the trade-off between granularity and the ability to handle long contexts. $\theta$ is often tuned to accommodate longer context lengths \citep{qwen2, roziere2023code}. Since our context length is significantly shorter than that of the original LLM, we explore reducing $\theta$ for potential performance gains. Our findings show that setting $\theta=10,000$ with a context length of $1024$ enhances performance, so we adopt this configuration moving forward. We note that since we increase the context length (from 512 to 1024), we need to reduce the batch size as well, to not run into memory problems when training. We reduce the batch size by a half and keep the same amount of gradient accumulation steps, which gives us an effective batch size of $128$. An ablation of this adaptation is provided in Appendix \ref{sub:ablation_context_bs}

\begin{figure}[t!]
  \includegraphics[width=\columnwidth]{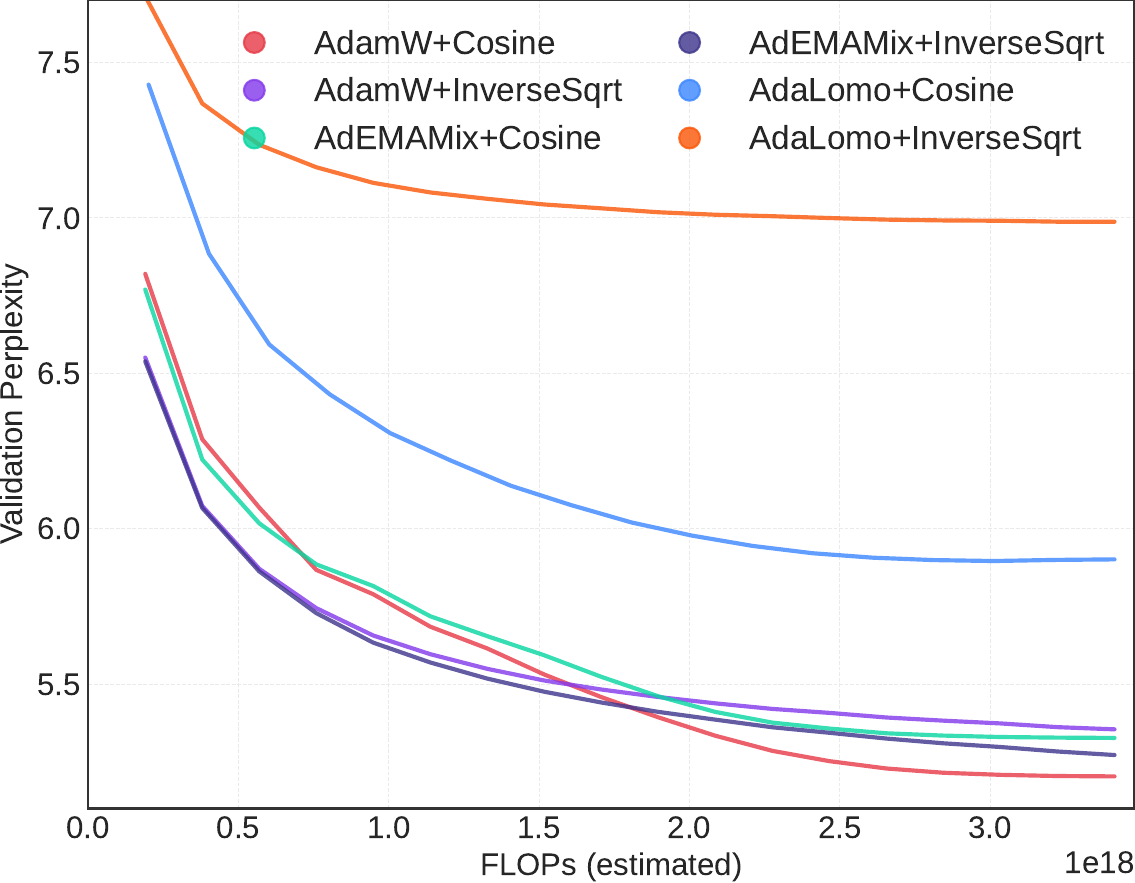}
  \caption{Comparing validation PPL of our best model with different optimisers and schedulers.}
  \label{fig:optim}
\end{figure}

\newpara{Optimiser and Scheduler.} Various optimisers and schedulers have been developed to enhance training efficiency, reduce memory usage \citep{shazeer2018adafactoradaptivelearningrates, dettmers20228bitoptimizersblockwisequantization}, or accelerate convergence \citep{pagliardini2024ademamix, chen2023lion}. With limited compute, these aspects become especially important. We first consider efficient optimisers, specifically AdamW with fused kernels, and $8$-bit AdamW, but observe no notable improvements in batch size or runtime compared to standard AdamW. This could do with the relatively small model size, resulting in a minimal memory footprint of the optimisers. We then compare AdamW with two state-of-the-art optimisers: AdaLomo \cite{lv2023adalomo} and AdEMAMeix \citep{pagliardini2024ademamix}. Results, presented in Figure~\ref{fig:optim}, suggest that with the original InverseSqrt scheduler used by \citet{twist}, using AdEMAMeix improves validation loss, compared to AdamW, with AdaLomo far behind.

Next, we analyse a cosine decay learning rate scheduler, in place of the original InverseSqrt as this was shown to improve convergence \cite{loshchilov2016sgdr}. We consider the previous optimisers, and provide the validation loss throughout training in Figure~\ref{fig:optim}. We see that this notably improved the loss for AdamW, and slightly harmed results for AdEMAMeix. Overall, AdamW with a cosine schedule provide the best setup, far outperforming the original setup.

\begin{figure}[t!]
  \centering
  \includegraphics[width=\columnwidth]{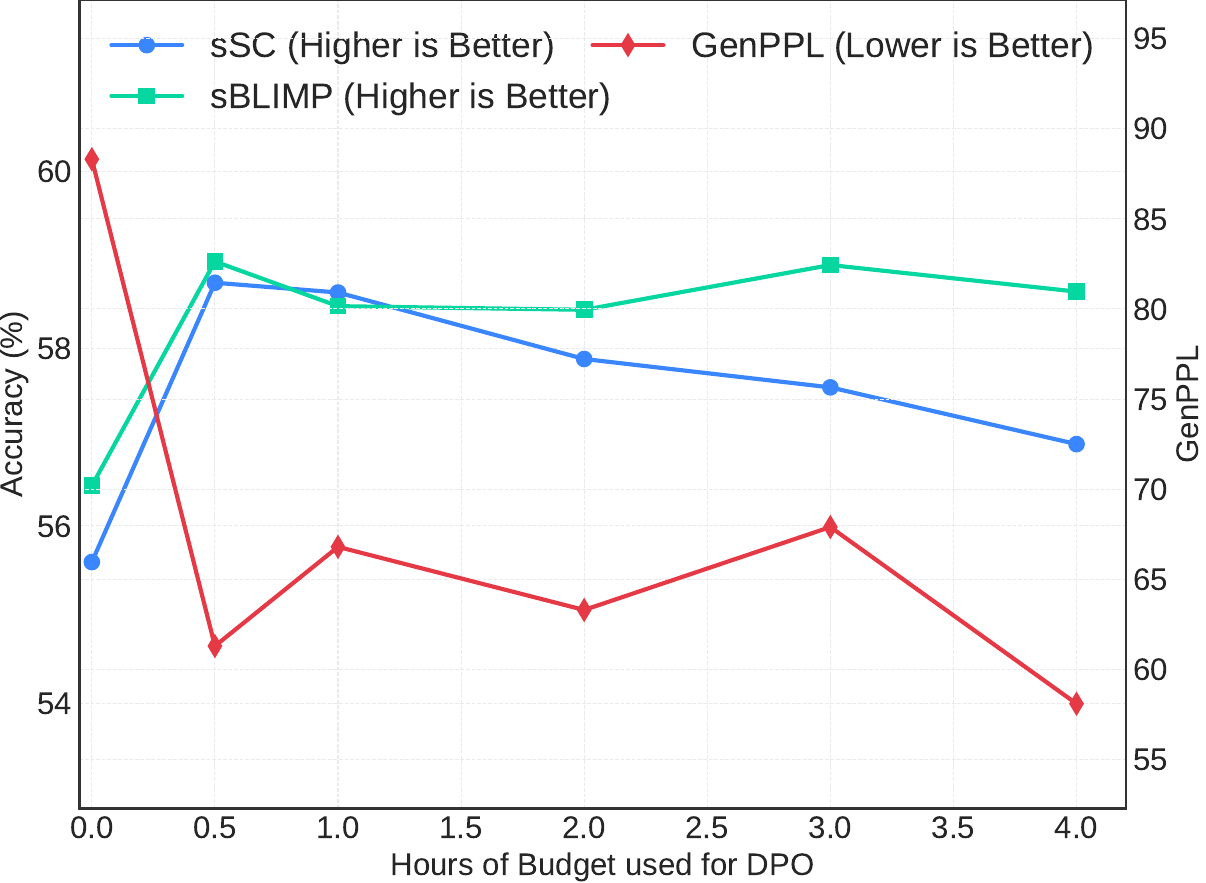}
  \caption{Analysing the optimal part of the 24 hour compute budget that should be used for DPO, with the rest used for pre-training.\label{fig:dpo}}  
\end{figure}

\subsection{Data} \label{sec:data}
Next, we examine how the training data-mix influences performance in a compute-constrained setting. Specifically, we explore whether diversity in accents, speaking styles, etc. is beneficial and assess if synthetic data can enhance semantic abilities. We provide exact statistics for each dataset in Appendix \ref{app:data_stats}.

\begin{table}[t!]
  \centering
  \resizebox{\columnwidth}{!}{
  \begin{tabular}{l|cc|cccc}
    \toprule
    Model & \multicolumn{2}{c|}{Data} & \multicolumn{4}{c}{Metric}\\
    \midrule
          & Div. & Syn. & sBLIMP$\uparrow$ & sSC$\uparrow$ & tSC$\uparrow$ & GenPPL$\downarrow$ \\
    \midrule
    OPT125M& \xmark & \cmark & 55.28 & \bf{55.46} & \bf{75.18} & \bf{96.8} \\
           & \cmark & \cmark & 55.06 & 55.00 & 74.83 & 116.6\\
           & \xmark & \xmark & \bf{55.88} & 54.52 & 70.82 & 160.3 \\
           & \cmark & \xmark & 55.65 & 54.78 & 70.18 & 172.7 \\           
    \midrule
    Qwen-0.5B & \xmark & \cmark & 56.45 & \bf{55.59} & \bf{78.01} & \bf{88.3} \\
              & \cmark & \cmark & 56.17 & 55.37 & 77.13 & 101.3 \\
              & \xmark & \xmark & \bf{56.60} & 53.50 & 71.14 & 145.4 \\
              & \cmark & \xmark & 56.10 & 53.72 & 70.66 & 161.8 \\
    \bottomrule
  \end{tabular}
  }
  \caption{Analysing impact of training data diversity and synthetic data on \slm performance. The default \method recipe does not use diverse data (only Libri-light and LibriSpeech), but uses the synthetic sTinyStories data.}
  \label{tab:data_analysis}
\end{table}

\newpara{Diverse Data.} We begin by examining how dataset diversity impacts model performance. Many leading speech datasets, such as those based on audiobooks \citep{ls, ll}, consist of relatively clean, single-speaker recordings within a specific content domain. To introduce greater diversity in speaking styles and content, we curate additional datasets, including VoxPopuli \cite{vp}, Tedlium \cite{tedlium}, PeopleSpeech \cite{people}, and SWC \cite{swc}. For all mentioned datasets, we use the official data cleaning and preprocessing scripts when available. Specifically, for Libri-light, we apply the official Voice Activity Detection model to remove silences and generate smaller audio segments. To evaluate the impact of dataset diversity, we compare the performance of \slms trained using our best training recipes using a subset of LibriSpeech and Libri-light against all curated datasets. This comparison is conducted for both OPT-$125$M, which processes a large number of tokens during training, and Qwen-$0.5$B, which encounters significantly less data due to model size. Results are summarised in Table~\ref{tab:data_analysis}. We observe that dataset diversity has an overall negative effect on model performance. We hypothesise this is due to the models struggling in modelling rich and complex audio under such low compute resources.

\begin{table*}[t!]
  \centering
  \resizebox{\textwidth}{!}{
  \begin{tabular}{l|ccccccc}
    \toprule
      & C (GPU days) & Params & sBLIMP$\uparrow$ & \ssc$\uparrow$ & \tsc$\uparrow$ & GenPPL$\downarrow$ & BLEU$\downarrow$\\

    \midrule
    TWIST-350M \cite{twist} & 40*V100 & 305M & 56.20 & $\emptyset$ & $\emptyset$ & 137.3 & 3.46 \\
    TWIST-1.3B \cite{twist} & 160*V100 & 1B & 57.00 & 52.4 & 70.6 & 131.8 & 3.20 \\
    TWIST-7B \cite{twist} & $\emptyset$ & 7B & 59.00 & 55.3 & 74.1 & 93.7 & 3.06 \\
    TWIST-13B \cite{twist} & $\emptyset$ & 13B & 59.20 & 55.4 & 76.4 & $\emptyset$ & $\emptyset$\\
    Scaled Optimal \cite{cuervo2024scaling} & $\emptyset$ & 823M & \bf{61.3} & \bf{56.7} & \bf{78.0} & $\emptyset$ & $\emptyset$\\
    \midrule
    Predicted Optimal \cite{cuervo2024scaling} & 1*A5000 & 78M & 56.85 & 54.09 & 70.49 & $\emptyset$ & $\emptyset$\\
    
    \midrule
    TWIST-350M (Original recipe) & 1*A5000 & 305M & 51.52 $\pm$.19 & 53.65 $\pm$.57 & 68.80 $\pm$.47 & 259.2 $\pm$6.7 &  3.26 $\pm$.46 \\
    TWIST-350M + sTinyStories & 1*A5000 & 305M & 51.21 $\pm$.26 & 54.17 $\pm$.54 & 72.40 $\pm$.18 & 159.0 $\pm$6.0 & 4.18 $\pm$.24 \\
    \method (-DPO) (ours) & 1*A5000 & 358M & \underline{56.45} $\pm$.17 & \underline{55.59} $\pm$.30 & \underline{78.01} $\pm$.27 & \underline{88.3} $\pm$1.0 & 3.47 $\pm$.17 \\
    \method (ours) & 1*A5000 & 358M & \textbf{58.86} $\pm$.20 & \textbf{58.04} $\pm$.51 & \textbf{82.04} $\pm$.21& \textbf{62.8} $\pm$4.1 & 3.88 $\pm$.11 \\
    \bottomrule
  \end{tabular}
}
\caption{Comparing \textit{slamming} to leading \slms, and predicted optimal performance for the compute. We also consider TWIST-$350$M using our code and compute budget, but with the original training recipe. $\pm$ indicates distance to min/max of $3$ seeds. BLEU is Auto-BLEU. \label{tab:slam}}
\end{table*}

\newpara{Synthetic Data.} Recent studies have highlighted the potential of synthetic data generated through \ac{TTS} \citep{cuervo2024scaling} or direct text-to-unit conversion \citep{scaling_interleaving}. Hence, we examine the impact of including synthetically generated speech within our constrained compute setup. To do so, we synthesised the TinyStories dataset \citep{tinystories} using a single-speaker \ac{TTS} model \citep{wang2021fairseq}, as it is computationally efficient. Additionally, prior research has shown that HuBERT units largely remove speaker information \citep{maimon2023dissc}. TinyStories has been demonstrated to enhance text \ac{LM} performance and improve \slms~\citep{cuervo2024scaling}. Results are presented in Table~\ref{tab:data_analysis}. Results indicate that incorporating such synthetic data into the training data-mix significantly boosts both modelling and generative performance metrics, across all evaluated setups. We also consider adding the synthetic data to the original TWIST recipe, and the results in the bottom of Table \ref{tab:slam} suggests that while this helps with semantic metrics, it is far from enough without other optimisations we introduced. As a further ablation, we assess the performance of \slm when trained exclusively on synthetic data. Results suggest, perhaps unsurprisingly, this leads to a significant drop in performance relative to our baseline model, which uses both real and synthetic data. Specifically, the model trained only on synthetic data scores $52.35$ on sBLIMP, compared to $56.45$ for the baseline, and exhibits a notably higher validation loss on real data ($2.8$ vs. $1.65$). We observe this across all datasets, and specifically with our best mixture Libri-Light, LibriSpeech and sTinyStories, Qwen-$0.5$B outperforms OPT-$125$M so we continue with it to the final stages. These findings reinforce the importance of incorporating both real and synthetic data during training.

\subsection{Text Interleaving} 
Several recent \slms combine both speech and text modalities, either predicting both simultaneously \citep{defossez2024moshi, fang2024llama, xie2024mini} or training on interleaved data \citep{spiritlm, scaling_interleaving}. Beyond enhancing cross-modal abilities, this has been shown to improve the semantic capabilities of \slms, even in speech-only evaluations. Building on these studies, we investigate whether speech-text interleaving can enhance semantic ability in speech-only tasks, even under strict computational constraints.

For this we use Whisper-large-v$3$-turbo to get aligned transcriptions of our data, except sTinyStories for which we get alignment from the \ac{TTS}. We follow \citet{scaling_interleaving} by selecting speech spans with length from a Poisson distribution with $\lambda=10$ totalling $30\%$ of the interleaved data. Following \citet{spiritlm} we train with balanced batches with respect to token count between text data, speech data and interleaved data. We use a subset of RedPajama \citep{weber2024redpajama} filtered by Gopher \citep{rae2021scaling} rules as our text data.

The \slm trained with interleaving with the exact same setup as the speech only variant slightly underperformed compared to the speech only. We report results as the mean of three training runs. Specifically, it achieved \tsc of 73.36 (compared to 78.01 for the speech only equivalent), \ssc of 55.76 (vs 55.59) and sBLIMP of 55.71 (vs 56.45). We note that the interleaved \slm has much larger vocabulary which in turn means that the model has more parameters ($\sim500M$ vs $~\sim360M$), which in turn means that each update step takes longer. For our budget the interleaved model only performed $\sim11k$ steps vs $\sim18k$ for speech only. Furthermore, out of all training tokens only about 40\% are speech tokens in the interleaved setting. This could perhaps explain the slightly worse performance, and we leave for future work to find the minimal compute budget to benefit from text-interleaving.

\begin{table*}[t!]
  \centering
  \resizebox{\textwidth}{!}{
  \begin{tabular}{l|ccccccccc}
    \toprule
      & GPUs & Params & Num tokens & sBLIMP$\uparrow$ & \ssc$\uparrow$ & \tsc$\uparrow$ & GenPPL$\downarrow$ & BLEU$\downarrow$ &GPTScore$\uparrow$\\
    \midrule
    \multicolumn{9}{l}{\bf{Speech only pre-training}} \\
    GSLM \cite{gslm}& 8*V100 & 100M & 1B & 54.2 & 53.3 & 66.6 & $\emptyset$ & $\emptyset$ & $\emptyset$\\
    SyllableLM \cite{baade2024syllablelm}& 4*A40 & 300M & 16B & 63.7 & $\emptyset$ & 75.4 & $\emptyset$ & $\emptyset$ & $\emptyset$\\
    TWIST-350M \cite{twist} & 8*V100 & 305M & 10.8B & 56.20 & $\emptyset$ & $\emptyset$ & 137.3 & 3.46 & $\emptyset$\\
    TWIST-1.3B \cite{twist} & 32*V100 & 1B & 10.8B & 57.00 & 52.4 & 70.6 & 131.8 & 3.20 & 1.82 \\
    TWIST-7B \cite{twist} & 32*V100 & 7B & 36B & 59.00 & 55.3 & 74.1 & 93.74 & 3.06 & 2.71\\
    TWIST-13B \cite{twist} & 32*V100 & 13B & 36B & 59.20 & 55.4 & 76.4 & $\emptyset$ & $\emptyset$ & $\emptyset$ \\
    \citet{cuervo2024scaling} & $\emptyset$ & 823M & 82B & \bf{61.3} & 56.7 & 78.0 & $\emptyset$ & $\emptyset$ & $\emptyset$\\
    Moshi \cite{defossez2024moshi} & ?*H100 & 7B & $\emptyset$ & 58.9 & \bf{58.7} & \bf{81.8} & $\emptyset$ & $\emptyset$ & $\emptyset$\\
    SpiritLM \cite{spiritlm} & 64*A100 & 7B & 100B & 58.0 & 54.8 & 72.9 & $\emptyset$ & $\emptyset$ & $\emptyset$ \\
    \midrule
    \multicolumn{9}{l}{\bf{Joint speech-text pre-training / preference optimisation}} \\
    \citet{scaling_interleaving} & $\emptyset$ & 9B & $\sim$1T & $\emptyset$ & \bf{62.4} & 82.9 & $\emptyset$ & $\emptyset$ & $\emptyset$\\
    Moshi \cite{defossez2024moshi} & ?*H100 & 7B & $\sim$720B & 58.8 & 60.8 & 83.0 & $\emptyset$ & $\emptyset$ & $\emptyset$\\
    SpiritLM \cite{spiritlm} & 64*A100 & 7B & 100B & 58.3 & 61.0 & 82.9 & $\emptyset$ & $\emptyset$ & $\emptyset$\\
    AlignSLM-1.3B \cite{lin2024alignslm} & 64*A100 & 1B & 10.8B + $\sim$158B & 59.8 & 55.0 & 80.0 & $\emptyset$ & $\emptyset$ & 2.43\\
    AlignSLM-7B \cite{lin2024alignslm} & 64*A100 & 7B & 36B + $\sim$158B & \bf{62.3} & 61.1 & \bf{86.8} & $\emptyset$ & $\emptyset$ & \bf{3.50}\\
    \midrule    
    \method (-DPO)       & 2*A100 & 358M & 16.7B & 58.53 & 58.15 & 80.71 & 67.3 & 3.25 & $\emptyset$ \\
    \method & 1*A5000 & 358M & 1.4B + 5M & 58.86 & 58.04 & 82.04 & 62.8 & 3.88 & 2.09\\        
    \method (scaled) & 2*A100 & 358M & 16.7B + 9M & 61.11 & 61.30 & 84.18 & 46.6 & 3.75 & 2.69 \\
    \method (large) & 2*A100 & 1.3B & 6.1B + 9M & \bf{61.43} & \bf{61.52} & \bf{85.30} & \bf{41.2} & 3.89 & \bf{2.79}\\
    \bottomrule
  \end{tabular}
}
\caption{Analysing the effect of scaling up compute for \method. \# tokens refers to total, not unique, tokens used for training (estimated from provided information). We separately mark DPO tokens with a +. BLEU is Auto-BLEU.}\label{tab:scaling}
\end{table*}

\subsection{Synthetic Data Preference Optimisation} 

Preference optimisation methods have been shown to enhance the performance of text LLMs \cite{ouyang2022training} and, more recently, \slms \cite{lin2024alignslm}. With preference optimisation, we aim to train our model to generate outputs that better align with a specified reward function or preference set.

We evaluate how preference optimisation affects \slm performance while considering our constrained computational budget. Using an \textit{off-policy} approach with pre-generated preference data, we apply DPO to enhance training efficiency. Specifically, we synthetically generate the SWAG \citep{swag} text corpus for evaluating semantic knowledge. SWAG consists of text prefixes paired with multiple possible suffixes, where only one is semantically plausible. For preference data, we use the first sentence as the prompt, the correct suffix as the positive continuation, and a randomly chosen incorrect suffix as the rejected continuation. To ensure quality, we filter out samples with repetitive patterns, identified by an auto-BLEU score above $0.3$. We generate all recordings using Kokoro TTS \citep{kokoro}, incorporating four speakers (two male and two female), evenly split between British and American accents. This process results in a total of $47$k SWAG preference pairs.  

For DPO we use $\beta=0.1$ (see Appendix \ref{sec:app_recipe} for full hyperparameters). In initial tests, we observe that after DPO training, the model shows increased likelihood at the cost of repeated patterns, a known issue with DPO~\citep{divpo}. To address this, we apply a repetition penalty with a factor of $1.1$, following the approach of \citet{ctrl}, and find that it helps mitigate the problem. Future work could explore alternative solutions, such as proposed by \citet{divpo}. 

We begin by examining how the allocation of budget for DPO impacts performance, particularly when it comes at the cost of a shorter pre-training phase. Figure~\ref{fig:dpo} depicts the results. We observe significant improvements across all metrics when applying DPO for at least $30$ minutes compared to not using DPO at all. However, allocating a higher proportion of the budget to DPO does not yield further gains and can even degrade model performance. Thus we stick to $30$ minutes out of $24$  hours for DPO, using the rest for pre-training.

\section{Final Recipe} 
\label{sec:recipe}
Building on these empirical findings, we develop the final \method recipe. Using it, we train \slms based on Qwen$2.5$-$0.5$B. We then compare \method to the TWIST model family across various sizes: $350$M, $1.3$B, $7$B, and $13$B. We also present results for TWIST-$350$M using our computational constraints but following TWIST's original training recipe, along with our synthetic data. Finally, we report results for the top-performing model from ~\cite{cuervo2024scaling}, including their predicted optimal performance under our compute budget based on \slm scaling laws. Results are reported in Table~\ref{tab:slam}. The results indicate that \method delivers performance that is either superior or on par with baseline models while requiring significantly fewer computational resources (e.g., a single A$5000$ for a day compared to $160$ days on a V$100$). Transcribed generated examples by \method can be seen in Appendix~\ref{sub:generation_examples}.

To show that the \method models do not overfit a single domain (audiobooks/stories), we provide results for GenPPL on a different domain. This can be seen in Appendix ~\ref{sub:different_domain_data}.

We further evaluate the quality of the generated audio using Mosnet~\cite{cooper2022generalization}, similarly to Align-SLM. Results are presented in Appendix~\ref{sub:mos_proxy_res}. As the quality of the generated audio is mainly affected by the vocoder, which is identical across evaluated methods, results are comparable. Interestingly, TWIST $1.3$B and TWIST $7$B achieve slightly worse scores.

\section{Increasing Compute} 
\label{sec:scale}

Similarly to \citet{geiping2023cramming}, we analyse whether the proposed approach holds well also in increased compute budget. We opt for $48$ hours on $2$ A$100$ GPUs as a reasonable academic budget for larger scale tests, and represents $\sim10$ times more compute than the \textit{Slamming} setting. We use exactly the same \method recipe for more steps, and increase the batch size times $2$. We provide the full results in Table \ref{tab:scaling}. We note that the performance continues to improve across all metrics, also outperforming methods which have far larger compute scales. We note that DPO training on synthetic data for $2$ epochs, notably boosts performance.  Transcribed generated examples by \method(scaled) can be seen in Appendix~\ref{sub:generation_examples}

We also wish to assess whether our suggested recipe holds for larger models, thus we evaluate training a larger Qwen$2.5$ text LM as the base model. We use Qwen$2.5-1.5$B for the same compute budget as above - i.e two A$100$ GPUs for $48$ hours. All training details are identical, but of course the larger model was trained for less steps (and tokens). We provide results from this model, denoted \method (large) in Table \ref{tab:scaling}. Results show that this model even outperforms the smaller model for the same compute budget. This demonstrates that the \method recipe holds for larger models, and re-iterates the importance of quality models even at the expense of less training tokens for this setup.

\section{Limitations}
While the \slms trained under \textit{Slamming} compute budget performed notably well compared to other \slms trained with much more compute they might perform less well in other areas. For instance, evaluating their abilities on acoustic or prosodic elements as in SALMon \cite{maimon2024salmon} could show further challenges of low resource settings.

Furthermore, we focus in this study on the well used HuBERT \cite{hubert} model as a tokeniser, and while we do not make any adjustments specifically for it, future work might wish to investigate our cramming approach with new tokenisers, such as Mimi \cite{defossez2024moshi} and SylBoost \cite{baade2024syllablelm}. 

\section{Conclusion}
In this work we show that training high quality \slms with a very modest compute budget, is feasible. We give these main guidelines: 
\begin{enumerate}
    \item \textbf{Do not skimp on the model} - not all model families are born equal and the TWIST initialisation exaggerates this, thus it is worth selecting a stronger / bigger text-LM even if it means less tokens. we found Qwen$2.5$ to be a good choice.
    \item \textbf{Utilise synthetic training data} - pre-training on data generated with TTS helps a lot.
    \item \textbf{Go beyond next token prediction} - we found that DPO boosts performance notably even when using synthetic data, and as little as $30$ minutes training massively improves results.
    \item \textbf{Optimise hyper-parameters} - as researchers we often dis-regard this stage, yet we found that tuning learning rate schedulers and optimising code efficiency can improve results notably.
\end{enumerate}

We hope that these insights, and open source resources will be of use to the community in furthering \slm research.

\section*{Ethical Statement}
The broader impact of this study is, as in any generative model, the development of a high quality and natural speech synthesis. We hope that allowing training \slms under low-resource settings, and open sourcing resources to aid this goal, will have a positive impact on inclusivity and accessibility of \slm research beyond well funded labs.

\paragraph{Acknowledgements.} This research work was supported by ISF grant 2049/22.

\bibliography{custom}
\clearpage
\appendix

\section{Full \method Recipe} \label{sec:app_recipe}
We provide below the full training recipe, including hyperparameters for the best, \method recipe. In Table~\ref{tab:slam-dpo-recipe} we see the \method (-DPO) pre-training recipe and in Table~\ref{tab:dpo-recipe} we see the \method DPO training recipe. Table~\ref{tab:slam-sampling} provides the sampling hyper-parameters used for calculating the generative metrics. Note that some of the generated samples in the demo page were created with a higher maximum token limit.

\begin{table}[ht]
  \centering
  \caption{\method(-DPO) pre-training recipe.}
  \resizebox{\columnwidth}{!}{
  \begin{tabular}{l|c}
    \toprule
    Parameter & Value \\ 
    \midrule
    Text Base Model   & Qwen2.5-0.5B  \\
    TWIST initialisation & True \\ 
    Data & Librilight, Librispeech, sTinyStories \\
    Train Time  & $23.5$ hours $\simeq17625$  steps  \\ 
    RoPE theta   & $10000$  \\ 
    Context length   & $1024$  \\ 
    Per device Batch Size   & $8$  \\
    Gradient Accumulation & $16$ \\ 
    Base Learning Rate & $1e-3$ \\ 
    Warmup Ratio & $1\%$ \\ 
    Optimizer   & AdamW  \\ 
    LR Scheduler   & cosine with min $5e-5$  \\
    Max Grad Norm & $0.5$ \\
    Dtype & bfloat16 \\
    \bottomrule
  \end{tabular}
  }
  \label{tab:slam-dpo-recipe}
\end{table}

\begin{table}[ht]
  \centering
  \caption{\method DPO training recipe.}
  \resizebox{\columnwidth}{!}{
  \begin{tabular}{l|c}
    \toprule
    Parameter & Value \\ 
    \midrule
    Initial Model & \method (-DPO) \\
    Data & SpokenSwag with auto-bleu$\leq0.3$ \\
    Train Time  & $0.5$ hour $\simeq813$  steps  \\ 
    RoPE theta   & $10000$  \\ 
    Context length   & $1024$  \\ 
    Per device Batch Size   & $4$  \\
    Gradient Accumulation & $16$ \\
    Base Learning Rate & $5e-5$ \\ 
    Optimizer   & AdamW  \\ 
    Learning Rate Scheduler   & inverse sqrt  \\
    Max Grad Norm & $0.5$ \\
    Dtype & bfloat16 \\
    DPO $\beta$ & $0.1$ \\
    \bottomrule
  \end{tabular}
  }
  \label{tab:dpo-recipe}
\end{table}

\begin{table}[ht]
  \centering
  \caption{\method sampling parameters.}
  \begin{tabular}{l|c}
    \toprule
    Parameter & Value \\ 
    \midrule
    Temperature & $0.8$ \\
    Top-K & $25$  \\
    Max New Tokens & $150$ \\
    Repetition Penalty & $1.1$ \\
    \bottomrule
  \end{tabular}
  \label{tab:slam-sampling}
\end{table}

\section{Model Sizes}\label{sec:model_sizes}

\begin{table}[h]
  \caption{Model names and parameter counts after changing vocabulary to speech only units (500).}
  \centering
  \resizebox{\columnwidth}{!}{
  \begin{tabular}{l|c}
    \toprule
    Model Name & Params\\
    \midrule
    MobileLLM-125M \cite{mobilellm} & 106,492,608 \\
    MobileLLM-350M \cite{mobilellm} & 315,117,120 \\
    OPT-125M \cite{opt} & 87,015,936 \\
    OPT-350M \cite{opt} & 305,714,176 \\
    QWEN2.5-0.5B \cite{qwen2} & 358,347,904 \\
    SmolLM2-135M \cite{smollm2} & 106,492,608 \\
    SmolLM2-360M \cite{smollm2} & 315,117,120 \\
    Pythia-160M \cite{pythia}  & 85,827,072  \\
    Pythia-410M \cite{pythia}  & 303,339,520  \\
    \bottomrule
  \end{tabular}
  \label{tab:model_sizes}
  }
\end{table}

As mentioned, we use the original names of the text LMs used for clarity and consistency, but note that the actual parameter counts after resizing the vocabulary to speech-units only can be very different. In Table \ref{tab:model_sizes} we provide an extensive list of models and sizes.

\section{Dataset Statistics}\label{app:data_stats}
We use and synthesise several datasets. In this section we give exact details of number of samples, splits used, domains etc. 

For pre-training we use Libri-Light \cite{ll} and LibriSpeech \cite{ls}. For Libri-Light we randonly select one percent of samples as validation, whereas for LibriSpeech we use the original \emph{dev-clean} and \emph{dev-other} splits. Both of these datasets are English speech only, focused in the audio-book domain. We also synthesise sTinyStories for pre-training which consists of synthetically generated English short stories. We use the official train split for training. Full dataset sizes are in Table \ref{tab:data_stats}.

We also investigate diverse datasets for pre-training: SWC \cite{swc}, Tedlium \cite{tedlium}, PeopleSpeech \cite{people} and VoxPopuli \cite{vp}. We only take English subsets for all datasets, yet they can still contain diverse accents. These datasets are in the following domains SWC - read Wikipedia articles, Tedlium - short lectures, PeopleSpeech - diverse data including many local council gatherings etc, VoxPopuli - from European Parliament meetings. For SWC specifically, we use the text alignment to create chunks, remove silence from the audio and remove mis-aligned chunks. We use full training splits where provided, otherwise splitting 99\% for training. The dataset sizes are described in Table \ref{tab:data_stats}.

\begin{table}[ht]
  \caption{Training set size for used datasets.}
  \centering
  \resizebox{\columnwidth}{!}{
  \begin{tabular}{l|cc}
    \toprule
    Dataset & Hours & Tokens\\
    \midrule
    Libri-Light \cite{ll} & $50K$ & $3.5B$ \\
    LibriSpeech \cite{ls} & $960$ & $67M$ \\
    SWC \cite{swc} & $750$ & $19M$ \\
    Tedlium \cite{tedlium} & $1.6K$ & $110M$ \\
    PeopleSpeech \cite{people} & $7K$ & $480M$ \\
    VoxPopuli \cite{vp} & $24K$ & $1.64B$ \\
    sTinyStories & $30K$ & $2.2B$ \\
    \bottomrule
  \end{tabular}
  \label{tab:data_stats}
  }
\end{table}

For DPO we synthesise SpokenSwag based on the SWAG \cite{swag} dataset. We use only the official train set and filter only the gold standard labels. We end up with 47k sample pairs which end up to be $\sim 4.5M$ tokens.

\section{Additional Results}

\subsection{Context Length and Batch Size Ablation}\label{sub:ablation_context_bs}

In Table \ref{tab:context_batch_ablation} we see results for ablations of context length and effective batch size

\begin{table}[h!]
    \caption{Performance on sBlimp, tStoryCloze (\tsc), sStoryCloze (\ssc) and validation loss across different context lengths and effective batch sizes. using Qwen2.5-0.5B}
    \centering
    \resizebox{\columnwidth}{!}{
        \begin{tabular}{ll|cccc}
        \toprule
        Context & BS & sBLIMP$\uparrow$ & tSC$\uparrow$& sSC$\uparrow$ & Val loss$\downarrow$ \\
        \midrule
        512 & 128   & 56.13 & 76.91 & 55.53 & 1.67 \\
        512 & 256   & \textbf{56.56} & 77.49 & \textbf{56.33} & 1.66 \\
        1024 & 256  & 56.43 & \textbf{78.88} & 55.69 & 1.65 \\
        1024 & 128  & 56.45 & 78.01 & 55.59 & \textbf{1.64} \\
        \bottomrule
        \end{tabular}
        \label{tab:context_batch_ablation}
    }
\end{table}

\subsection{MOS Proxy Results}\label{sub:mos_proxy_res}

For completeness we also provide MOS proxy results for our models compared to TWIST and Align-SLM models. We follow a similar setup to \citet{lin2024alignslm} and use MOSnet to test the audio's generation quality of our models. It is important to note that we use the same vocoder as TWIST and Align-SLM. The results can be seen in Table~\ref{tab:mosnet_scores}.

\begin{table}[h!]
    \caption{MOSnet scores for various models.}
    \centering
    \resizebox{\columnwidth}{!}{
    \begin{tabular}{l|c}
    \toprule
    Model & MOSnet $\uparrow$ \\
    \midrule
    Ground Truth    & 4.28 \\
    \midrule
    TWIST-350M \citep{twist} & 4.07 \\
    TWIST-1.3B \citep{twist} & 3.83 \\
    TWIST-7B   \citep{twist} & 3.85 \\
    Align-SLM-1.3B \citep{lin2024alignslm} & 4.05 \\
    Align-SLM-7B   \citep{lin2024alignslm} & 4.09 \\
    \method         & \textbf{4.11} \\
    \method(scaled) & 4.07 \\
    \method(large)  & 4.05 \\
    \bottomrule
    \end{tabular}
    \label{tab:mosnet_scores}
    }
\end{table}

\subsection{GenPPL on Different Domain}\label{sub:different_domain_data}

In order to evaluate the generalisability of our approach to diverse domains, we calculate GenPPL for a dataset from a different domain. We compare our results to TWIST of various sizes, which were trained on this exact dataset (perhaps even overlapping samples as no official training set was published). We use the same setup for GenPPL as described in section~\ref{para:metrics}, but we use People Speech test set~\cite{people} as prompts. Results in Table~\ref{tab:gen_ppl_ps_scores}, show that \method performs comparably or better to TWIST models of larger scale and more train compute, despite the fact they were explicitly trained on this dataset. These results highlight the efficacy of the \method recipe beyond a single domain. 

\begin{table}[h!]
    \caption{GenPPL results on the People Speech Dataset.}
    \centering
    \resizebox{\columnwidth}{!}{
    \begin{tabular}{l|cc}
    \toprule
    Model & GenPPL $\downarrow$ & BLEU $\downarrow$\\
    \midrule
    TWIST-350M \citep{twist}     & 162.89 & 2.36 \\
    TWIST-1.3B \citep{twist}     & 155.51 & 2.86 \\
    TWIST-7B   \citep{twist}     & 128.74 & 2.72 \\
    \method         & 171.84 & 4.89 \\
    \method(scaled) & 128.78 & 3.99 \\
    \method(large)  & \textbf{117.49} & 3.72 \\
    \bottomrule
    \end{tabular}
    \label{tab:gen_ppl_ps_scores}
    }
\end{table}

\subsection{Text Generation Examples}\label{sub:generation_examples}

For completeness we provide some transcriptions for some of the generated examples (generated by \method and \method(Scaled)). The prompts and the transcriptions of the generated samples can be seen in Table \ref{tab:slam_examples}.

\begin{table*}[t!]
\caption{transcriptions of generated continuations from Slam and Slam Scaled for various prompts.}
\centering
\renewcommand{\arraystretch}{1.3}
\begin{tabular}{p{2.5cm}|p{6.0cm}|p{6.5cm}}
\toprule
Prompt & Slam & Slam Scaled \\
\midrule
Johnny found an old coin on the subway floor… &
and decided to bring it home. Then he came up by a different subway floor, which opened into a courtyard. Johnny picked up the coin and threw it at the doorstep of the courtyard. &
He picked it up and looked at the size and shape of the coin. It must be valuable, remembering his promise to Johnny, had it put very carefully in his pocket. But now he felt as if somebody had given him a special coin in return. \\
\hline
I am a real boy he said &
He looked at Sarah Weinfeild and smiled. It was quite impossible for Sarah to look at him any longer, and had no suspicions about his feelings for a moment. &
So off he went to school in the spring time, sitting on the soft grass near the trees where the rain fell in torrents. As the sun began to set, the birds sang their lovely songs and the children clapped their hands with Joe and Jack Pensilvenya. \\
\hline
I had a great day today playing &
I saw a little boy playing too. He said hello to me, I went to say hello to you. I said hello back. I thought he was very nice, I asked him if he wanted to play with me? He said yes, and they played together telling jokes and laughter in the quiet little world. They had a great time together, and when it was time to go home I said bye to the little boy and went to sleep. end. &
I love playing the flute, and almost forgot my troubles in the warm sunshine as I set the flute for miles around in the air practicing my favorite songs. My flute singing out in answer to the beautiful sounds from the world around me. At length, the song ended and I stood up and clapped my hands, then we passed on through the forest, over the meadow and along the river bank until we came to a beautiful valley. \\
\bottomrule
\end{tabular}
\label{tab:slam_examples}
\end{table*}

\section{AI Tool Usage}
AI based tools may have been used in writing parts of the code for this study, or para-phrasing some of the writing within the paper, yet all the content was thoroughly checked by the authors, with these only being used as assistive tools.

\end{document}